\documentclass[a4paper,11pt]{article}

\usepackage{algorithm}
\usepackage{algpseudocode}
\usepackage{amsmath}
\usepackage{amssymb}
\usepackage{amsthm}
\usepackage{graphicx}
\usepackage{authblk}

\newcommand{\beq}{\begin{equation}}
\newcommand{\eeq}{\end{equation}}

\newcommand{\ba}{\begin{array}}
\newcommand{\ea}{\end{array}}

\newcommand{\bea}{\begin{eqnarray}}
\newcommand{\eea}{\end{eqnarray}}

\newcommand{\bc}{\begin{center}}
\newcommand{\ec}{\end{center}}

\newcommand{\bt}{\begin{tabular}}

\newcommand{\bi}{\begin{itemize}}
\newcommand{\ei}{\end{itemize}}

\newcommand{\bd}{\begin{description}}
\newcommand{\ed}{\end{description}}

\newcommand{\bp}{\begin{pmatrix}}
\newcommand{\ep}{\end{pmatrix}}

\newcommand{\p}{\partial}

\newcommand{\cf}{{\it cf.}~}
\newcommand{\gnorm}[1]{\left|\left|#1\right|\right|}

\title{ \bf Entropic Regression DMD (ERDMD) Discovers Informative Sparse and Nonuniformly Time Delayed  Models}

\author[1]{Christopher W. Curtis}
\author[1,2]{Daniel Jay Alford-Lago}
\author[3,4]{Erik Bollt}
\affil[1]{Department of Mathematics and Statistics, SDSU}
\affil[2]{Naval Information Warfare Center}
\affil[3]{Department of Electrical and Computer Engineering, Clarkson University}
\affil[4]{Clarkson Center for Complex Systems Science, Clarkson University}
\date{}
\begin{document}
\maketitle

\begin{abstract}
In this work, we present a method which determines optimal multi-step dynamic mode decomposition (DMD) models via entropic regression, which is a nonlinear information flow detection algorithm.  
Motivated by the higher-order DMD (HODMD) method of \cite{clainche}, and the entropic regression (ER) technique for network detection and model construction found in \cite{bollt, bollt2}, we develop a method that we call ERDMD that produces high fidelity time-delay DMD models that allow for nonuniform time space, and the time spacing is discovered by consider most informativity based on ER. These models are shown to be highly efficient and robust.  We test our method over several data sets generated by chaotic attractors and show that we are able to build excellent reconstructions using relatively minimal models.  We likewise are able to better identify multiscale features via our models which enhances the utility of dynamic mode decomposition.   
\end{abstract}

\section{Introduction}

Equation free model development has enjoyed a multi-year stretch of continued progress in part by way of the evolution of dynamic mode decomposition (DMD) methods.  Beginning in the fluid dynamics community (see \cite{taira} for a historical review and key sources), DMD has moved into almost every area of data driven science, and it has merged itself with every major trend in the data sciences as well, in particular machine learning based methods; see \cite{lusch, azencot, lago_dldmd}.  

Likewise, in order to better address the modeling of multiscale data, delay structured DMD models reminiscent of either ARIMA or Takens embedding models have also been explored; see \cite{arbabi, clainche, duraisamy, champion2, kutz4, curtis_dldmd}.  Across these works, it has become clear that the success of DMD when used to generate models which can be time stepped forward from some set of initial data depends upon finding accurate multi-step Koopman like models for a given time series.  However, much of the existing literature has in practice focused on data coming from periodic and quasi-periodic attractors, which as is proved in \cite{duraisamy}, can be exactly reconstructed using relatively straightforward lagged DMD models.  While \cite{arbabi} shows that Hankel-DMD methods should provide exact determination of the affiliated Koopman operator in ergodic systems, this is not necessarily a practical result.  This issue is partially addressed in \cite{curtis_dldmd}, which coupled an adaptive Hankel DMD method to an autoencoding neural network strategy to discover meaningful observables from chaotic time series.  Nevertheless, this method did not provide a fully automatic way to determine the updating of the Takens style embeddings needed to make the method successful.  Thus, discovering methods which build better lagged DMD models has a number of potential downstream applications.  

We treat discovering lagged DMD models as a question of optimal model discovery.  While there are a plethora of approaches, the one that we find most impactful is \cite{bollt2}, which treats the problem of finding the most likely model in terms of the flow of information between time scales.  Information flow is measured here via a generalization of transfer entropy \cite{schreiber}, called {\it causation entropy} \cite{bollt}, coupled with an algorithmic approach to model discovery called {\it entropic regression} \cite{bollt2}, making our approach one which seeks to find those models which provide the most information about evolving time series.  Our work then is related to methods which build inferential causal networks in multidimensional time series \cite{idtxl2}, and it is also a natural compliment of \cite{lozano}, which uses similar concepts to transfer entropy to identify causal coupling across scales in turbulent fluids.  

As we show, our method, which we dub the entropic-regression-dynamic-mode decomposition (ERDMD), is able to discover non-uniform lagged DMD models which provide accurate time-stepping schemes from data coming from a variety of chaotic attractors.  The ability of our method to generate non-uniform lagged models echoes the work in \cite{vlachos, faes}, which shows non-uniform embeddings of time series lead to more information theoretically rich models.  In all then, the present work realizes the ambition of the results in \cite{arbabi} while pointing towards future improvements in the algorithm presented in \cite{curtis_dldmd}.  That said, we also find limitations by way of our results on the Kuramoto--Sivashinsky equation, which shows even optimally lagged DMD models struggle in of themselves to accurately reconstruct higher dimensional strange attractors.  Future work in this direction by way of coupling our current method with the work of \cite{curtis_dldmd} should address this shortcoming.  Likewise, as noted in \cite{bollt2}, a major motivation for using information theory in model discovery is that it is robust to noisy data.

The structure of the paper is as follows.  In Section 2, we provide a brief introduction to higher order DMD and entropic regression.  We then present our central algorithm, the ERDMD method.  In Section 3, we present the results of our method across three different chaotic dynamical systems.  Finally in Section 4, we provide a discussion of the method and further elaborate on future directions which would follow from the present work.   

\section{Entropic Regression DMD}

The results in this work are a merging of the higher-order DMD (HODMD) method of \cite{clainche}, and the entropic regression technique for network detection and model construction found in \cite{bollt, bollt2}.  We now briefly explain both results, and then we show how to bring them together in order to build accurate, yet minimal time lag models.     

\subsection{Higher-Order DMD}

Originally presented in \cite{clainche}, though see also the HAVOK method in \cite{brunton_havok} and Hankel DMD method of \cite{arbabi}, we generalize the HODMD so as to keep each lagged model separate, causal, and formulated over an arbitrary sequence of not necessarily unit increment lags.  We suppose that we have the maximum lag of $d$ time steps.  We likewise suppose that we have the choice of say $N_{l}$ lags as $l_{c}=\left\{l_{1}, l_{2}, \cdots, l_{k}, \cdots, l_{N_{l}}\right\}$, with $1=l_{1} < l_{j} < l_{j+1} < l_{N_{L}}<d$.  Given times series $\left\{{\bf y}_{j}\right\}_{j=0}^{N_{T}}$, we then define the matrices 
\[
{\bf Y}_{+,d} = \left({\bf y}_{d} \cdots {\bf y}_{N_{T}}\right), ~ {\bf Y}_{-} = \left({\bf y}_{0} \cdots {\bf y}_{N_{T}-1}\right), ~ {\bf y}_{j}\in\mathbb{R}^{s},
\]  
and the $s\times (N_{T}-d+1)$ shifted mask matrices ${\bf M}_{l}$ where
\[
\left({\bf M}_{l}\right)_{mn} = \left\{
\ba{rl}
0 & m-1 < d-l, ~ m-1 > N_{T} - l\\
1 & n=m, ~ d-l\leq m-1 \leq N_{T} - l\\
0 & n\neq m, ~ d-l\leq m-1 \leq N_{T} - l
\ea
\right.
\]

An arbitrary lagged DMD model can then be written as 
\begin{equation}
{\bf Y}_{+,d} = \sum_{k=1}^{N_{L}}{\bf K}_{l_{k}}{\bf Y}_{-}{\bf M}_{l_{k}},
\label{basic_model}
\end{equation}
where we note that each matrix ${\bf K}_{l_{k}}$ is $s\times s$.  Using our standard optimization arguments, we can find each matrix ${\bf K}_{l_{j}}$ via the critical-point equation
\[
\sum_{k=1}^{N_{L}}{\bf K}_{l_{k}}{\bf Y}_{-}{\bf M}_{l_{k}}{\bf M}_{l_{j}}^{T}{\bf Y}_{-}^{T} = {\bf Y}_{+,d}{\bf M}_{l_{j}}^{T}{\bf Y}_{-}^{T}.
\]
Combining these terms across lags leads to the system 
\[
{\bf K}(l_{c}) {\bf Y}_{-}(l_{c}){\bf Y}_{-}(l_{c})^{T} = {\bf Y}_{+, d}{\bf Y}_{-}(l_{c})^{T}.
\]
where
\[
{\bf K}(l_{c}) = \left({\bf K}_{l_{N_{L}}} {\bf K}_{l_{N_{L}-1}} \cdots {\bf K}_{1}\right), 
\]
and
\[
{\bf Y}_{-}(l_{c}) = \begin{pmatrix} {\bf Y}_{-}{\bf M}_{l_{N_{L}}} \\ {\bf Y}_{-}{\bf M}_{l_{N_{L}}-1} \\ \vdots \\ {\bf Y}_{-}{\bf M}_{1} \end{pmatrix}.
\]

Once the model in Equation \eqref{basic_model} is built, we run it by coupling outputs to inputs so that for $j\geq d$, we have 
\begin{equation}
{\bf y}_{j+1} = \sum_{k=1}^{N_{L}}{\bf K}_{l_{k}}{\bf y}_{j+1-l_{k}}.  
\label{iterator}
\end{equation}
While the method sees data within the original given time series, we say the method is performing {\it reconstruction}.  When it begins to iterate over data it has not seen, then we say the model is performing {\it forecasting}. 

In \cite{clainche}, models with distinct matrices ${\bf K}_{l_{j}}$ for each lag were eschewed for something much closer to what is now called Hankel DMD.  While effective, and in several respects simpler, such an approach does not as readily allow for more thoughtful model selection as we explain in the next section.  That said, a price is paid for working with separate lag matrices when one wants to look at the corresponding one-step Koopman operator, say $\mathcal{K}^{\delta}$ which would be approximated by
\begin{equation}
\mathcal{K}^{\delta} \approx {\bf K}_{a} \equiv 
\begin{pmatrix} 
0 & I & 0 & \cdots & 0 & 0 \\ 
0 & 0 & I & \ddots & 0 & 0 \\ 
\vdots & \vdots & \vdots & \ddots & \ddots & \vdots \\
0 & 0 & 0 & \cdots & I & 0 \\
{\bf K}_{l_{N_{L}}} & \tilde{{\bf K}}_{l_{N_{L}}-1} & \cdots & \cdots & \tilde{{\bf K}}_{2} & {\bf K}_{1} 
\end{pmatrix}
\label{full_matrix}
\end{equation}
where each $s\times s$ matrix $\tilde{{\bf K}}_{j}$ is given by 
\[
\tilde{{\bf K}}_{j} = \left\{
\ba{rl}  
{\bf K}_{l_{k}}, & j=l_{k}\in l_{c}\\
{\bf 0}, & j\neq l_{k}\in l_{c}
\ea
\right.
\]
Finding the eigenvalues of ${\bf K}_{a}$ is equivalent to finding the roots of the polynomial $p_{a}(z)$ where 
\begin{equation}
p_{a}(z) = \text{det}\left(\sum_{k=0}^{N_{L}-1}{\bf K}_{l_{N_{L}-k}}~z^{l_{N_{L}}-l_{N_{L}-k}} - z^{l_{N_{L}}}{\bf I} \right). 
\label{characpoly}
\end{equation}
While not useful in a numerical context, this formula will prove useful when characterizing the eigenvalues of ${\bf K}_{a}$ relative to the choice of lags $l_{c}$.  

\subsection{Determining Information Flow and Model Discovery through Entropic Regression}

Given two time series, say $\left\{X_{j}\right\}_{j=1}^{N_{T}}$ and $\left\{Y_{j}\right\}_{j=1}^{N_{T}}$, it is a foundational question to determine if one time series {\it causes} the other.  Said another way, can we find quantitative methods which determine how one time series might drive or ultimately explain the behavior of another?  

Motivated by the now celebrated {\it Granger causality} test, \cf \cite{granger}, in linear time series, \cite{schreiber} introduced the notion of {\it transfer entropy} to determine the causal relationship between two nonlinear time series.  The transfer entropy from $X_{j}$ to $Y_{j}$, say $T_{X\rightarrow Y}(j)$ is defined in \cite{schreiber} to be 
\begin{align}
T_{X\rightarrow Y}(j) = & H\left(Y_{j+1}|Y_{j}\right) - H\left(Y_{j+1}|Y_{j}, X_{j}\right),  \\
= & I\left(Y_{j+1}, X_{j}|Y_{j} \right).
\end{align}
where $H()$ measures the entropy of a random variable and $I(,)$ is the mutual information between two random variables defined via the equation
\begin{equation}
I(X,Y) = H(X) - H(X|Y).  
\end{equation}
Thus the transfer entropy $I(Y_{j+1},X_{j}|Y_{j})$ measures the conditional mutual information with conditioning over $Y_{j}$.  Note, if $Y_{j+1}$ is independent of $X_{j}$, then $H(Y_{j+1}|Y_{j},X_{j}) = H(Y_{j+1}|Y_{j})$ so that $T_{X\rightarrow Y}(j) = 0$.  This initial concept of transfer entropy has given rise to a host of modifications and improvements, the most relevant for our purposes being that of {\it causation entropy} \cite{bollt} and {\it entropic regression} (ER) \cite{bollt2}.  Causation entropy provides a natural extension of transfer entropy to networks in which a central question is the determination of information flow between network nodes.  Central to its computation over a network is the separation of the algorithm, after some INITIALIZE phase, into a BUILD and PRUNE phase.  The PRUNE phase in particular removes nodes which would otherwise provide false-positive connections for information flow in the network, thereby providing a critical mechanism for developing accurate pictures of information flow among several processes.  

Building on this three stage methodology, ER develops models of time series from dictionaries of basis functions, say $\left\{\mathbf{\phi}_{l}\right\}_{l}^{N_{m}}$.  This happens as well in an  INITIALIZE, BUILD, and then PRUNE phase.  After choosing some reasonable initial basis function and corresponding model, the BUILD phase is done by fixing a state, say ${\bf Y}_{s}$, which is modeled by the current model, say ${\bf M}_{c}$, and a test model, say ${\bf M}^{(j)}_{t}$, where ${\bf M}^{(j)}_{t}$ differs from ${\bf M}_{c}$ through the inclusion of a basis function, say $\mathbf{\phi}_{j}$, not already used in constructing ${\bf M}_{c}$.  We then find a potential model update by computing
\[
\tilde{I} = \text{argmax}_{j}I\left(\left.{\bf Y}_{s}, {\bf M}_{t}^{(j)}\right| {\bf M}_{c}\right),
\]
so that, if $\tilde{I}>0$ in a statistically significant sense, we choose those model updates which most increase the degree of information that one model provides over another regards to a state ${\bf Y}_{s}$.  Once we have exhausted the family of basis functions, ER commences the PRUNE phase, in which for the current model ${\bf M}_{c}$ we generate ${\bf M}_{t}^{(j)}$ by removing basis functions, say $\phi_{j}$, from ${\bf M}_{c}$, and then compute 
\[
\tilde{i} = \text{argmin}_{j}I\left(\left.{\bf Y}_{s}, {\bf M}_{t}^{(j)}\right| {\bf M}_{c}\right),
\]
so that if $\tilde{i}=0$ in a statistically significant way, then we reduce ${\bf M}_{c}$ to the chosen ${\bf M}_{t}^{(j)}$.  
\subsection{Entropic-Regression-Dynamic-Mode Decomposition}
\begin{figure}
\centering
\includegraphics[scale=.75]{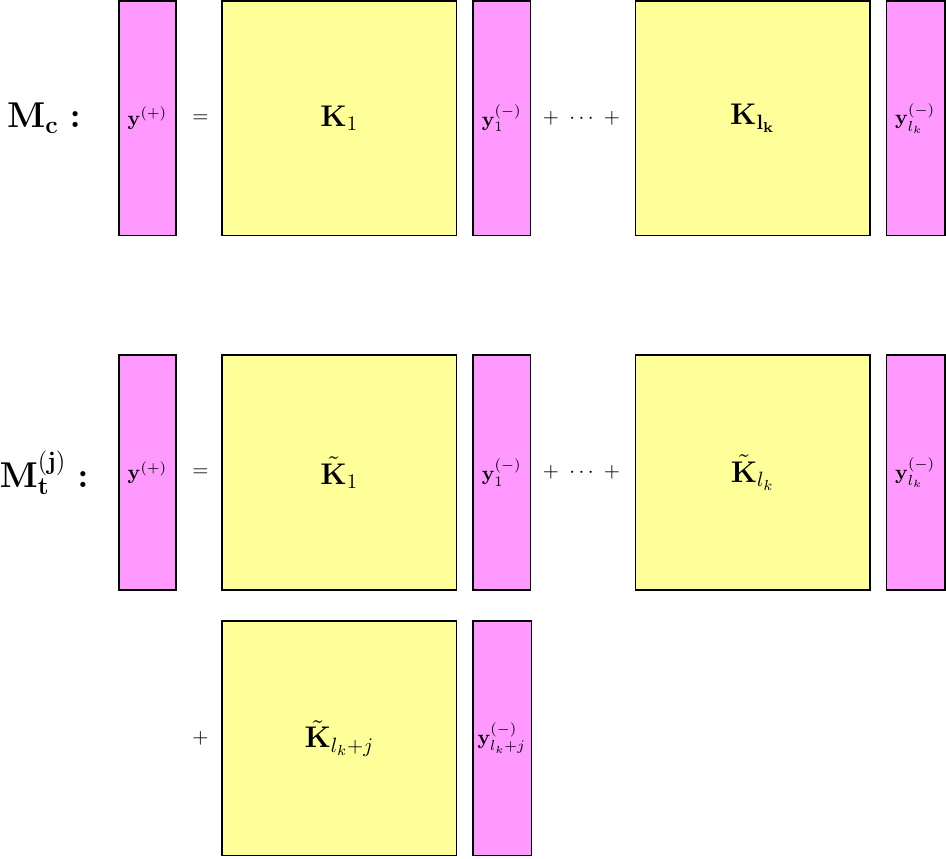}
\caption{The BUILD stage of the ERDMD algorithm.  Here, we start from the given model ${\bf M}_{c}$ which represents the choice of lags $(1, \cdots, l_{k})$ with corresponding matrices $({\bf K}_{1}, \cdots, {\bf K}_{l_{k}})$, and we then find a proposed model ${\bf M}^{(j)}_{t}$ which maximizes $I\left(\left.{\bf y}^{+}, {\bf M}^{(j)}_{t}\right|{\bf M}_{c}\right)$, or the information gain in using the proposed model relative to the given model to anticipate the next time steps represented by ${\bf y}^{(+)}$.}  
\label{fig:build_stage}
\end{figure}

To augment the HODMD method, we now merge it with ER.  To do this, fixing some maximum choice of lag $d$, we take as our desired state ${\bf Y}_{s}\equiv {\bf Y}_{d}^{+}$.  Assuming some already chosen set of lags, say $l_{c}$, we can add to $l_{c}$ some not already chosen lag , say $l_{j}$, to generate the set of test lags $l_{t}$.  We can then use HODMD to generate a given model ${\bf M}_{c}\equiv {\bf K}(l_{c}){\bf Y}(l_{t})$ and a proposed model ${\bf M}_{t}^{(j)}\equiv {\bf K}(l_{t}){\bf Y}(l_{t})$.  Following then the ER framework, there are three stages to our ERDMD method, which are 
\begin{enumerate}
\item {\bf INITIALIZE}: We initialize our choice of lags $l_{c}=\left\{1\right\}$ and corresponding HODMD matrix ${\bf K}_{1}$ which is found via the typical DMD method.    
\item {\bf BUILD}: Given some choice of lags $l_{c}$, we then look at the proposed list $l_{t}=l_{c}\cup \left\{l_{j}\right\}$ and then find those lagged models, say ${\bf M}^{(j)}_{t}\equiv {\bf K}(l_{t}){\bf Y}_{-}(l_{t})$ which provide the most information relative to the prior choice of lagged model, say ${\bf M}_{c}\equiv {\bf K}(l_{c}){\bf Y}_{-}(l_{t})$, i.e. we find 
\begin{equation}
\tilde{I} = \text{argmax}_{j}I\left(\left.{\bf Y}^{+}_{d},{\bf K}(l_{t}){\bf Y}_{-}(l_{t})\right|{\bf K}(l_{c}){\bf Y}_{-}(l_{t})\right).  
\label{build_eqn}
\end{equation}
If this value is larger than zero in a statistically significant way (measured through shuffle testing), then the model is updated to the corresponding model ${\bf M}^{(j)}_{t}$; see Figure \ref{fig:build_stage} for an illustration of this process.    
\item {\bf PRUNE}: We finally prune by testing whether each chosen lag is necessary relative to the other choices we have made during the build phase.  Thus, for a given model ${\bf M}_{c}\equiv {\bf K}(l_{c}){\bf Y}_{-}(l_{t})$, we generate ${\bf M}^{(j)}_{t}\equiv {\bf K}(l_{c}\backslash l_{j}){\bf Y}_{-}(l_{t})$ and then find 
\begin{equation}
\tilde{i} = \text{argmin}_{j}I\left(\left.{\bf Y}^{+}_{d},{\bf M}^{(j)}_{t}\right|{\bf M}_{c}\right).  
\label{prune_eqn}
\end{equation}
If this value is close to zero in a statistically significant way (measured through shuffle testing), then model is updated to the corresponding ${\bf M}^{(j)}_{t}$.  
\end{enumerate}
Note, we always have $1\in l_{c}$ since this makes all subsequent lags improvements on the basic DMD approach.  This process is formalized and detailed in Algorithm \ref{erdmd}.  
\begin{algorithm}
\caption{ERDMD Method}
\begin{algorithmic}[1]
    \Procedure{Initialize}{}
        \State Set $l_{c}=\left\{1\right\}$ and $l_{r}=\left\{2, \cdots, d\right\}$.
        \State Find ${\bf K}_{1} =  \text{arg min}_{\bf K}\gnorm{{\bf Y}^{+}_{1}-{\bf K}{\bf Y}_{-}\left(l_{c}\right)}_{F}$.  Set ${\bf K}(l_{c})=({\bf K}_{1})$.
    \EndProcedure
\Procedure{Build}{}
		\While{$l_{r}\neq \left\{\varnothing\right\}$}
        \State Given $l_{c}=\left\{1~l_{1} \cdots l_{j}\right\}$, ${\bf K}({\bf l}_{c})$, and $l_{r}=\left\{2,\cdots,d\right\}\backslash l_{c}$        
        \For{$l_{j+1}\in l_{r}$}
        \State Define $l_{t}=l_{c}\cup \left\{l_{j+1}\right\}$ and find
        \[
        {\bf K}(l_{t}) = \text{arg min}_{\bf \tilde{K}_{1}, {\bf \tilde{K}}_{l_{1}}, \cdots {\bf \tilde{K}}_{l_{j+1}}}\gnorm{{\bf Y}^{+}_{l_{j+1}}-{\bf \tilde{K}}(l_{t}){\bf Y}_{-}(l_{t}) }_{F}.
        \]          
    \EndFor
    \State Choose $l_{j+1}$ and the corresponding $l_{t}$ and ${\bf K}(l_{t})$ to maximize
        \[
        I\left(\left. {\bf Y}^{+}_{d}, {\bf K}(l_{t}){\bf Y}_{-}(l_{t})\right|{\bf K}(l_{c}){\bf Y}_{-}(l_{t})\right)
        \]
        \State If choice is statistically significant (using shuffle test), update $l_{c}$ and ${\bf K}(l_{c})$.
    \EndWhile
    \EndProcedure
    \Procedure{Prune}{}
        \State Given $l_{c}=\left\{1,l_{1}, \cdots, l_{N_{L}}\right\}$, set $S \equiv \text{True}$
        \While{$S$}
        \For{$l_{j}\in l_{c}$}
        	\State Define $l_{t}=\left\{1,l_{1}, \cdots, l_{N_{L}}\right\}\backslash \left\{l_{j}\right\}$
            \State Compute
            \[
            I\left(\left. {\bf Y}^{+}_{l_{N_{L}}}, {\bf K}(l_{t}){\bf Y}_{-}(l_{t})\right|{\bf K}(l_{c}){\bf Y}_{-}(l_{t})\right).
            \]
            
        \EndFor
        \State Choose $l_{t}$ corresponding to minimum information.  
        \If{minimum is statistically insignificant}
        \State Prune corresponding $l_{j}$ from $l_{c}$ and ${\bf K}(l_{c})$.
        \Else
        \State $S\equiv \text{False}$
        \EndIf
        \EndWhile
    \EndProcedure
\end{algorithmic}
\label{erdmd}
\end{algorithm}

\section{Results}

To study the efficacy of our method, we examine its use over numerically generated data from various chaotic dynamical systems.  In each case, for a given time series $\left\{{\bf y}_{j}\right\}_{j=0}^{N_{T}}$ we choose a maximum lag $d$ and then use a derived model to reconstruct the original time series after the $d^{th}$ step, i.e. $\left\{{\bf y}_{j}\right\}_{j=d}^{N_{T}}$. In each case, we compare the model generated by our proposed ERDMD method to the full HODMD method.  While this is of course not exhaustive with regards to model performance comparison, we refer the reader to \cite{curtis_dldmd}, which looks at the performance of the Hankel DMD method on a similar class of problems where we see it generally fail outright without further modification.  Thus, within its appropriate method class, we have a clear sense that the ERDMD method is performing well across its most nearby competitors.  
\subsection{Lorenz-63}
We now examine how well our method performs on the Lorenz-63 system, given by the equations
\begin{align*}
\dot{y}_{1} & = \sigma(y_{2}-y_{1})\\
\dot{y}_{2} & = y_{1}(\rho - y_{3}) - y_{2}\\
\dot{y}_{3} & = y_{1}y_{2}-\beta y_{3}\\
\end{align*}
with $\sigma=10$, $\rho=28$, and $\beta = 8/3$.  These parameter choices ensure that trajectories are pulled onto a strange attractor and exhibit chaotic dynamics.  We test our method on numerically generated data.  Time stepping is done with standard Runge--Kutta 4.For our numerics, the time step is $dt=.01$, and we run the simulation from $0\leq t \leq 22$.  

\begin{figure}[!h]
\centering
\begin{tabular}{c}
\includegraphics[width=.8\textwidth]{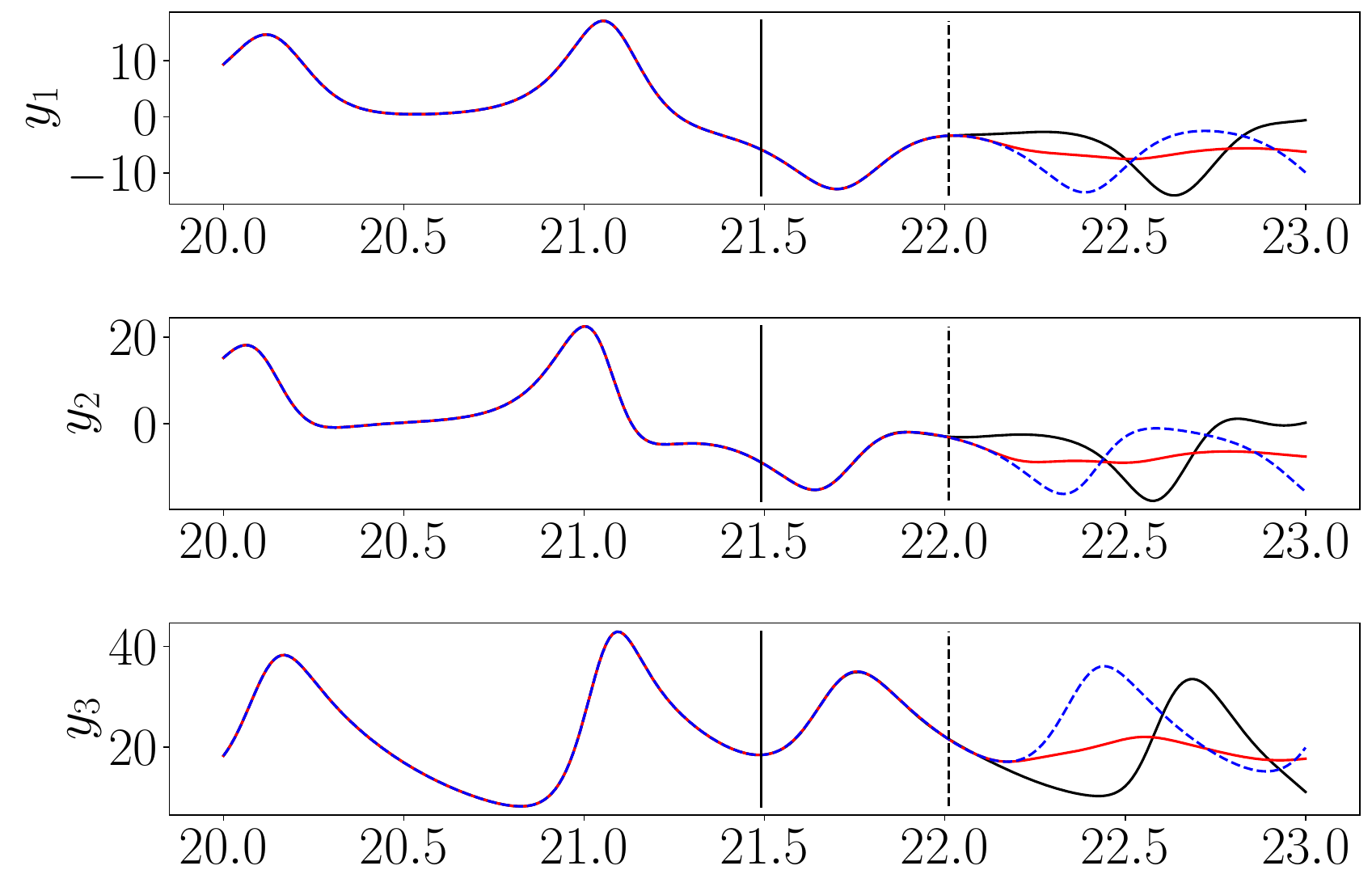}\\
(a) Trajectories\\
\includegraphics[width=.8\textwidth]{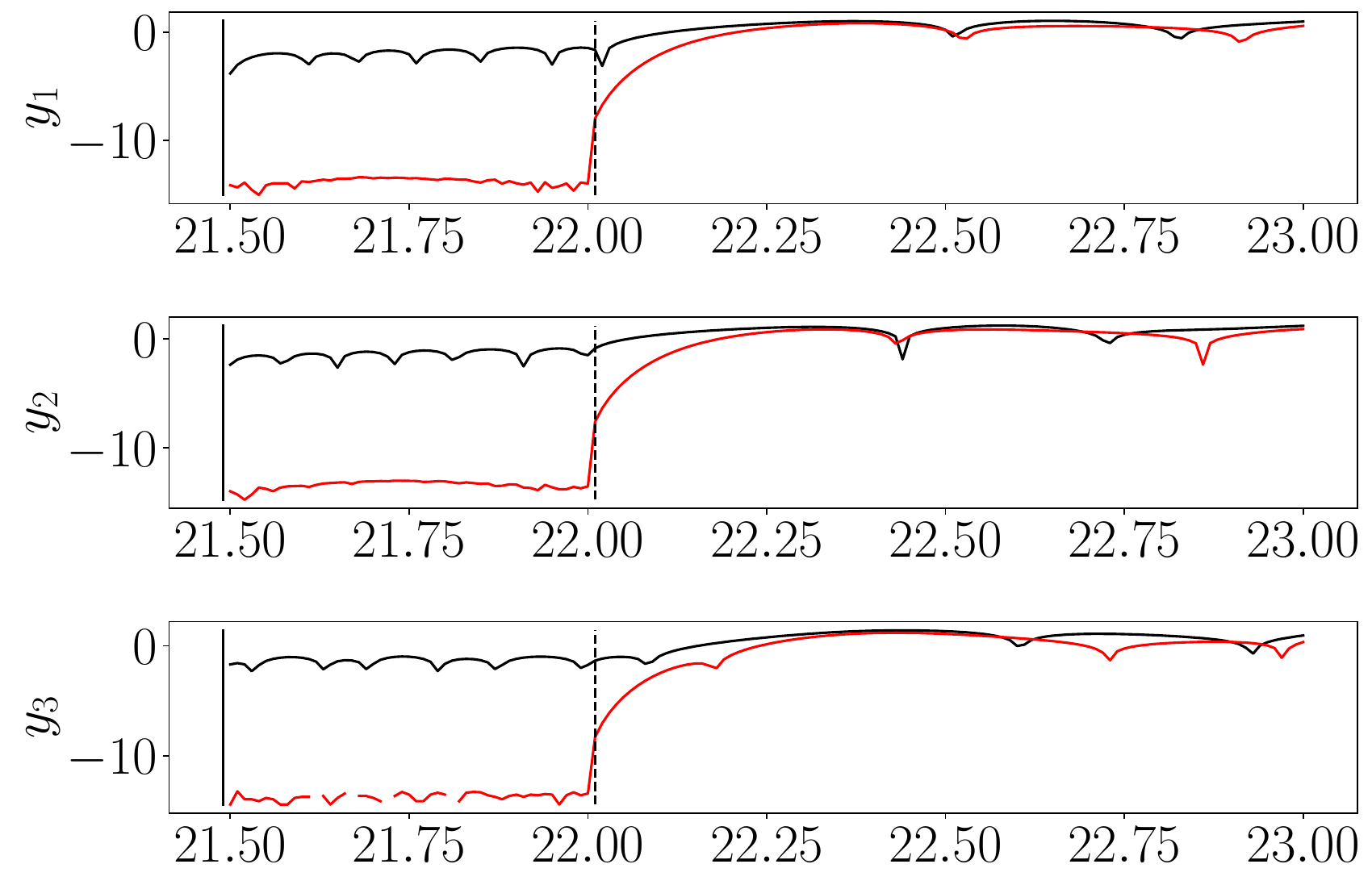}\\
(b) Error Comparison (semi-log scale)
\end{tabular}
\caption{Direct comparison of ERDMD and all lags HODMD model against the true trajectory for the Lorenz-63 system (a), and error across dimensions for the ERDMD and all lags HODMD method (b).  The black line indicates the ERDMD result while the red indicates the all lags HODMD result.  The solid vertical bar indicates the maximum lag choice of $d=150$, while the dashed line indicates the end of the reconstruction interval and the beginning of the forecasting regime. The ERDMD algorithm converges to $l_{c}=\left\{1,149\right\}$.}
\label{fig:lorenz_compare_d_150}
\end{figure}

We look at data for $20\leq t \leq 22$, with $d=150$, corresponding to $1.5$ units of non-dimensional time.  We then compare our ER-DMD model to a model using all possible lags for $21.5\leq t \leq 22$.  The ERDMD algorithm converges to $l_{c}=\left\{1,149\right\}$.  The results of running the two models in both reconstruction and forecasting modes are seen in Figure \ref{fig:lorenz_compare_d_150}.  In the error plots in Figure \ref{fig:lorenz_compare_d_150} (b), we see that throughout the reconstruction region, the HODMD is categorically more accurate than the ERDMD method by several orders of magnitude.  However, practically speaking, the approximation provided by the ERDMD is still quite good.  Further, looking beyond the reconstruction to the forecast regime, we see the ERDMD model does a generally better job beyond the data, thereby implying that the HODMD model is probably overfitting.  

\begin{figure}[!h]
\centering
\includegraphics[width=.7\textwidth]{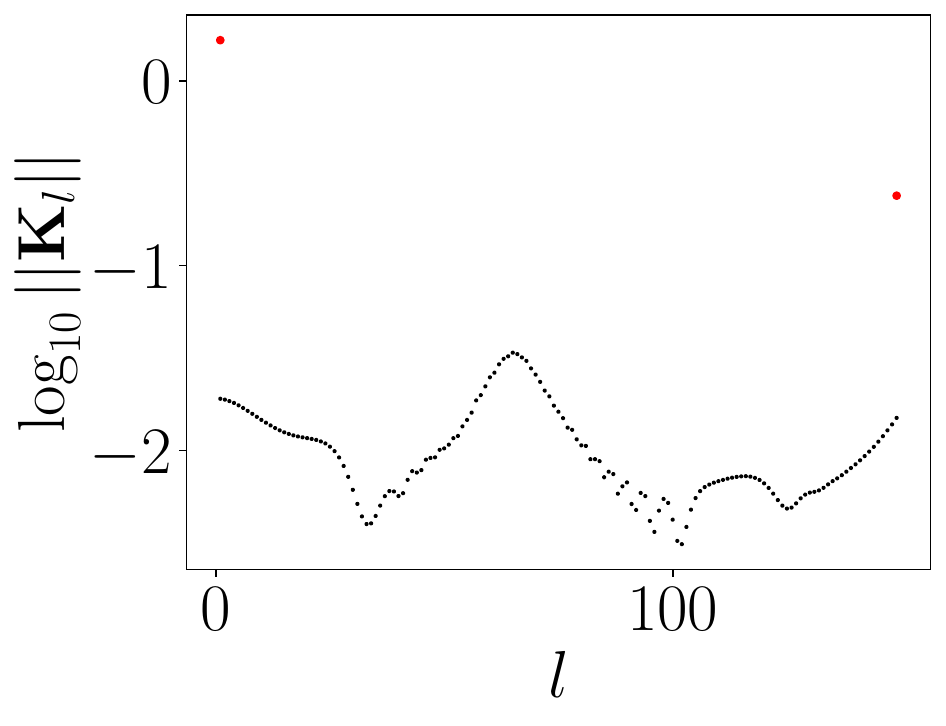}
\caption{Comparison of the HODMD lagged matrix norms (black dots) and the ERDMD model (red dots) for the Lorenz-63 system with $d=150$.}
\label{fig:model_comp_d_150}
\end{figure}

To this end, we can also compare the models generated by ERDMD and HODMD.  Looking at the norms of the various lag matrices in Figure \ref{fig:model_comp_d_150} we see that the ERDMD model produces a much sparser model with significantly larger overall matrix norms.  Likewise, the relative continuity of magnitudes of the lag matrices of the HODMD method allows for some description of the preferred scales of the model, but it is nowhere near as clear as when using ERDMD to accomplish the same task.  

\begin{figure}[!h]
\centering
\begin{tabular}{c}
\includegraphics[width=1\textwidth]{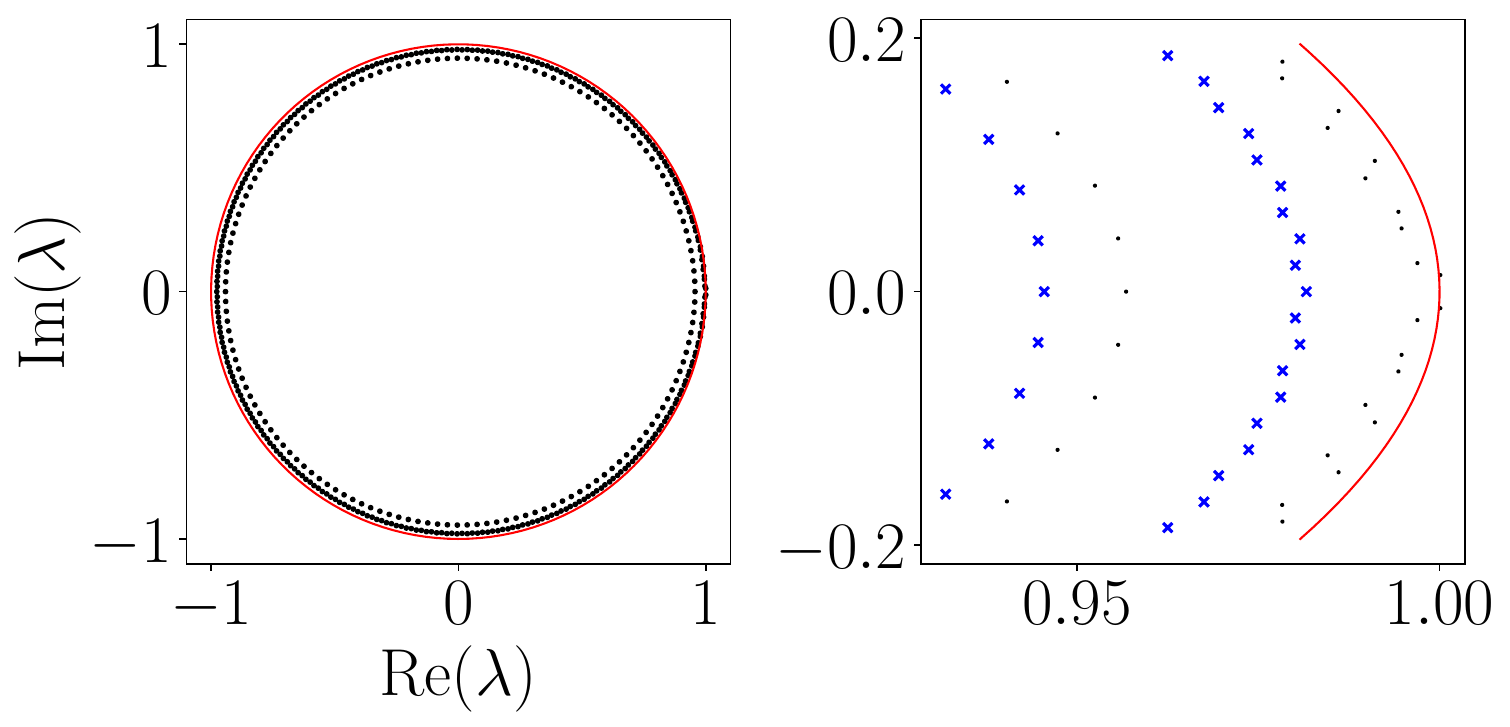}
\end{tabular}
\caption{Spectrum of corresponding ERDMD Koopman operator for Lorenz-63 with $d=150$ on the left side, with a detail comparison to the roots of $\tilde{p}_{in, a}(z)$ (blue crosses) on the right side near (1,0).  The ERDMD algorithm converges to $l_{c}=\left\{1,149\right\}$.  The solid/red line is the unit circle, provided for reference.}
\label{fig:lorenz_spectrum_d_150}
\end{figure}

Constructing the full Koopman one-step approximation ${\bf K}_{a}$ as in Equation \eqref{full_matrix} allows us to find the affiliated Koopman spectrum as seen in the left side of Figure \ref{fig:lorenz_spectrum_d_150}.  We see that the affiliated characteristic polynomial $p_{a}(z)$ is given explicitly by
\[
p_{a}(z) = \text{det}\left({\bf K}_{149} + \left({\bf K}_{1}-z{\bf I}\right)z^{148} \right).
\]
If we look at the interior of the unit disc so that $|z|<1$, then because of the strong separation in lags as seen in $l_{c}$, our innermost eigenvalues to leading order are found from the roots of $\tilde{p}_{in, a}(z)$ where
\[
\tilde{p}_{in, a}(z) = \text{det}\left({\bf K}_{149} + {\bf K}_{1}z^{148}\right)
\]
or the leading roots are found by finding the generalized eigenvalues $\tilde{z}=z^{148}$ of the two matrix problem ${\bf K}_{149} + {\bf K}_{1}\tilde{z}$.  

Looking at the detail figure in Figure \ref{fig:lorenz_spectrum_d_150}, we see that most of the features in the spectrum seen in the full spectrum on the left are present in the right.  Thus, the damping in the dynamics comes almost entirely from the disparity in lag values.  Further, we see that the full spectrum gets closer to the unit circle and even has two modes which just cross the unit circle.  Otherwise, the reduced model well describes the damping modes, though we also see that the lag matrices ${\bf K}_{1}$ and ${\bf K}_{149}$ balance for more delicate dynamics along the unit circle.    

\begin{figure}[!h]
\centering
\includegraphics[width=.8\textwidth]{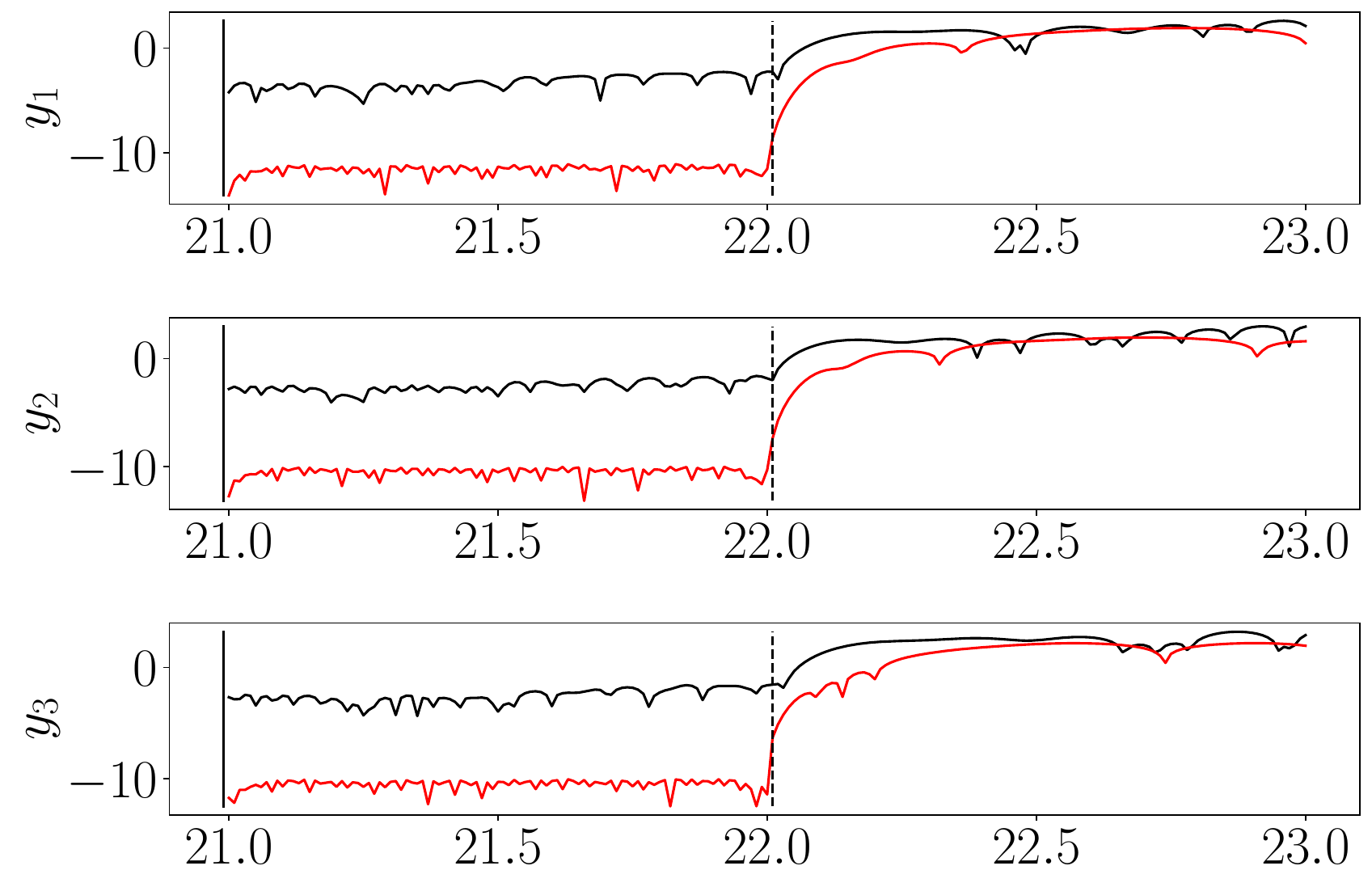}
\caption{Comparison of ERDMD and all lags HODMD model against true trajectory for Lorenz-63 system.  The ERDMD reconstruction is in solid black in the figure.  The vertical bar indicates the maximum lag choice of $d=100$. The ERDMD algorithm converges to $l_{c}=\left\{1,15, 26, 35, 45, 48, 68, 73, 97,99\right\}$.}
\label{fig:lorenz_compare_d_100}
\end{figure}

By way of contrast though, if we set the maximum lag $d=100$, and build reconstructions for $21\leq t \leq 22$, we find that the ERDMD results differ  markedly in terms of the determined lags though ultimately not in terms of accuracy; see Figure \ref{fig:lorenz_compare_d_100}.  In this case, the ERDMD algorithm converges onto the lag choices 
$$
l_{c}=\left\{1,15, 26, 35, 45, 48, 68, 73, 97,99\right\}.
$$  
By choosing a lag horizon which does not fully capture the approximate period of oscillation in the dynamics, we need significantly more information to accurately reconstruct the data, though still nowhere as much as the full HODMD model.  We also see though that the smaller choice of $d$ causes both models to essentially fail at providing any reasonable forecast, again reflecting the importance of good choices of $d$ in the first place, a result which echoes the work in \cite{duraisamy}.
 
\begin{figure}[!h]
\centering
\includegraphics[width=.7\textwidth]{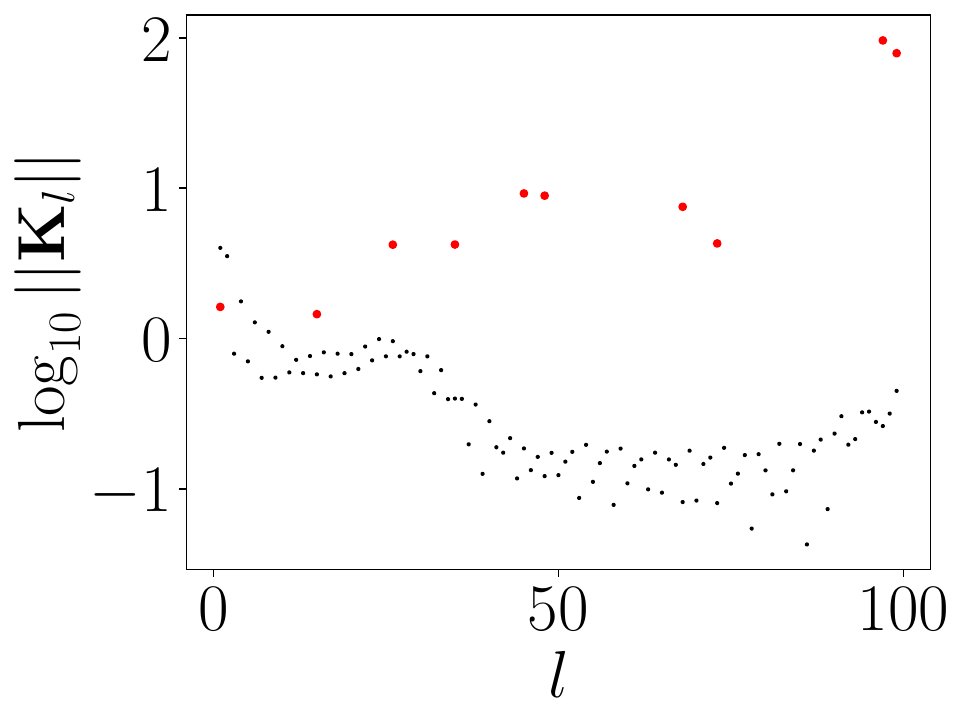}
\caption{Comparison of the HODMD lagged matrix norms (black dots) and the ERDMD model (red dots) for the Lorenz-63 system with $d=100$.}
\label{fig:model_comp_d_99}
\end{figure}

Comparing the models again as seen in Figure \ref{fig:model_comp_d_99}, we see that for $d=100$ we get a less minimal ERDMD model though with markedly larger matrix norms for the longer lags, helping us to see that the model prioritizes long correlations in time in order to find accurate reconstructions.  Likewise, the full HODMD model has some degree of spread in magnitudes, but it does not allow for ready analysis or description.    
 
\begin{figure}[!h]
\centering
\includegraphics[width=.5\textwidth]{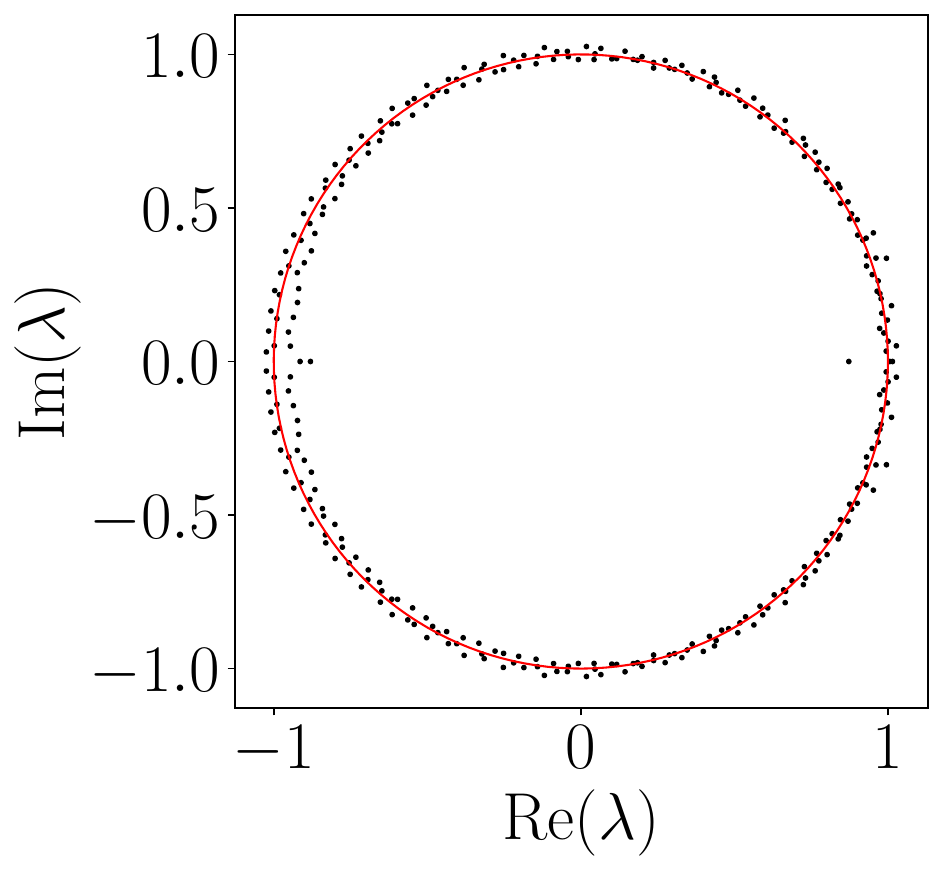}
\caption{Spectrum of corresponding ERDMD Koopman operator for Lorenz-63 for $d=100$ on the left side, with a detail comparison to the reduced approximation (blue crosses) on the right side near (1,0).  The ERDMD algorithm converges to $l_{c}=\left\{1,15, 26, 35, 45, 48, 68, 73, 97,99\right\}$.  The solid/red line is the unit circle, provided for comparison.}
\label{fig:lorenz_spectrum_d_100}
\end{figure}

We can likewise find the spectrum of ${\bf K}_{a}$ as seen in Figure \ref{fig:lorenz_spectrum_d_100}.  In this case, the larger spread of lags makes ready identification of positions in the spectrum to lag structure more difficult, though we see that 
The more uniform spread in lag values makes any estimates of the spectrum using reduced models less useful beyond identifying the most strongly damped modes.  

\subsection*{Rossler Equation}

To see how the ERDMD method works on problems with multiple scales, we now look at modeling dynamics coming from the Rossler system, given by the equations
\begin{align*}
\dot{y}_{1} & = -y_{2} - y_{3}\\
\dot{y}_{2} & = y_{1} + ay_{2}\\
\dot{y}_{3} & = b + y_{3}(y_{1}-c)\\
\end{align*} 
where $a=.1$, $b=.1$, $c=14$.  To see the role the multiple time scales play in this problem, letting $\epsilon=.1$, then we see $a=b=\epsilon$ and $c=1.4/\epsilon$.  Letting $\tau = t/\epsilon$ and setting $y_{3} = \epsilon^{2}\tilde{y}_{3}(t,\tau)$, to leading order, in $y_{1}$ and $y_{2}$ we find 
\[
\begin{pmatrix}y_{1} \\ y_{2}\end{pmatrix} = e^{\epsilon t/2}\begin{pmatrix}\cos(t) & -\sin(t) \\ \sin(t) & \cos(t)\end{pmatrix}\begin{pmatrix}y_{1,0} \\ y_{2,0}\end{pmatrix} + \mathcal{O}(\epsilon^{2}),
\]
so that we have $\mathcal{O}(1)$ planar oscillations complimented by slow growth away from the origin.  Likewise, in $\tilde{y}_{3}$ we find 
\[
\p_{\tau}\tilde{y}_{3} + \epsilon \p_{t}\tilde{y}_{3} = 1 - 1.4\tilde{y}_{3} + \epsilon y_{1}\tilde{y}_{3}.
\]
This then motivates the expansion 
\[
\tilde{y}_{3}(t,\tau) = c(t) e^{-1.4\tau} + \frac{1}{1.4}\left(1-e^{-1.4\tau} \right) + \epsilon \tilde{y}_{3,1}(t,\tau) + \cdots, 
\]
which, to remove secularities then gets us the equation for $c(t)$ of the form
\[
\frac{dc}{dt} = y_{1}(t)\left(c - \frac{1}{1.4} \right).
\]
So we have 
\[
c(t) = \frac{1}{1.4} + \left(c(0)-\frac{1}{1.4}\right)e^{\int_{0}^{t}y_{1}(s)ds},
\]
so that $\tilde{y}_{3}$ stays small until the slow growth in $y_{1}$ due to the $e^{\epsilon t/2}$ term pushes the dynamics out of the plane.  

\begin{figure}[!h]
\centering
\begin{tabular}{c}
\includegraphics[width=.8\textwidth]{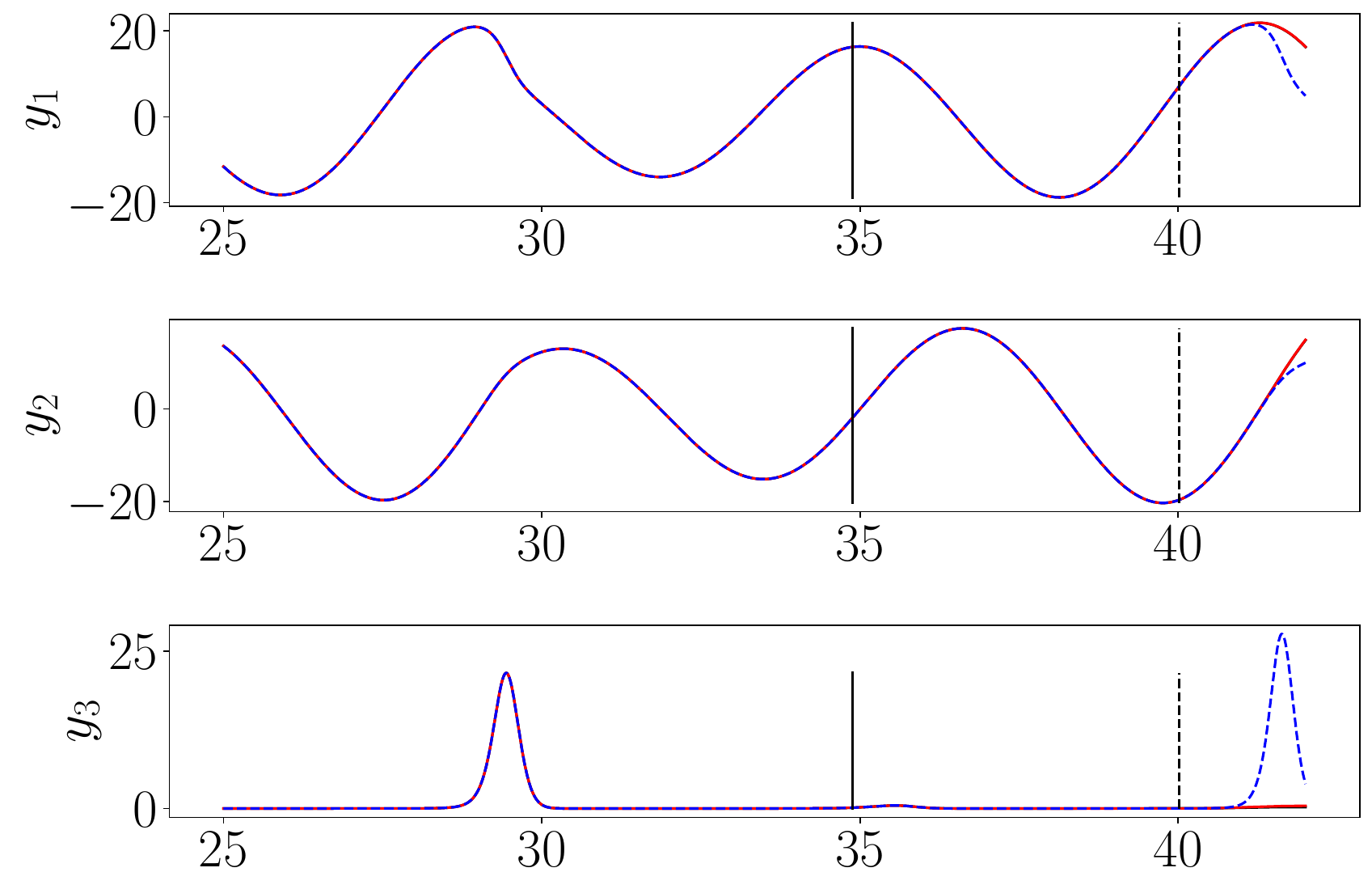}\\
(a) Trajectories \\
\includegraphics[width=.8\textwidth]{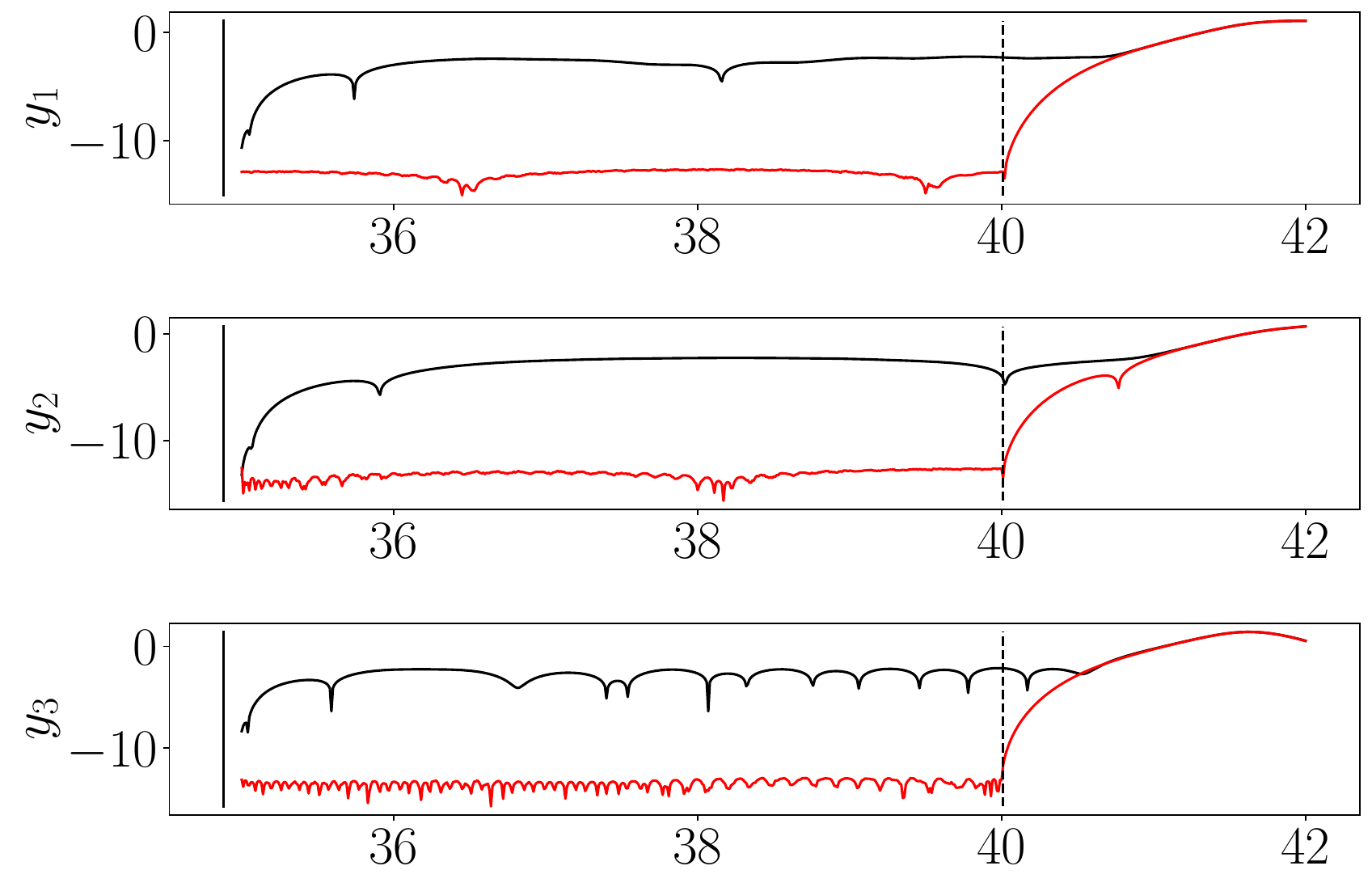}\\
(b) Error Comparison
\end{tabular}
\caption{Comparison of ERDMD and all lags HODMD model against true trajectory for the Rossler system.  The ERDMD reconstruction is in solid black in the figure.  The vertical bar indicates the maximum lag choice of $d=1000$. The ERDMD algorithm converges to $l_{c}=\left\{1, 3, 170, 436, 553, 665, 988\right\}$.}
\label{fig:rossler}
\end{figure}

We build reconstructions for $25\leq t \leq 40$.  To get accurate results, we chose $d=1000$, for which choice the method converged to the set $l_{c}$ with
\[
l_{c} = \left\{1, 3, 170, 436, 553, 665, 988\right\}.
\]
As can be seen in Figure \ref{fig:rossler}, our accuracy relies on capturing a full departure from the fast through the slow manifold.  This also explains the need for such a large choice of $d$ relative to what was used for the Lorenz-63 system.  Likewise, as with the Lorenz-63 system, the error in the ERDMD approximation is several orders of magnitude larger than for HODMD, though the practical difference is minimal.  Neither model does well forecasting the next fast departure from the planar oscillator reflecting an inherent difficult in learning slow/fast chaotic dynamics.  

\begin{figure}[!h]
\centering
\includegraphics[width=.7\textwidth]{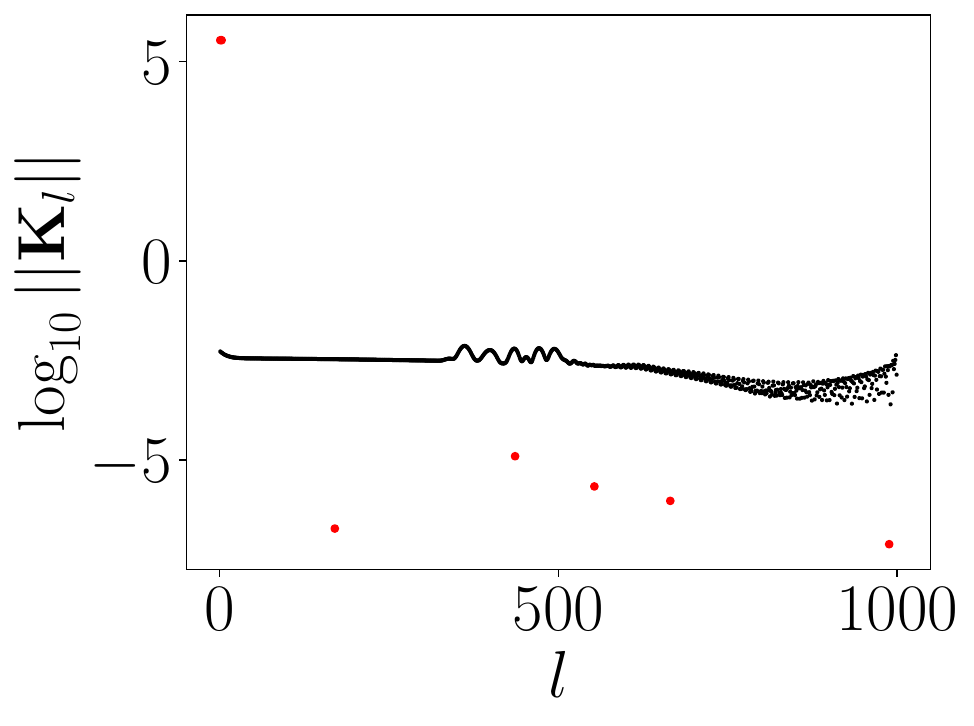}
\caption{Comparison of the HODMD lagged matrix norms (black dots) and the ERDMD model (red dots) for the Rossler system.}
\label{fig:model_comp_d_988}
\end{figure}

Comparing the HODMD and ERDMD models for the Rossler system, we see in Figure \ref{fig:model_comp_d_988} somewhat peculiar results.  Unlike for the Lorenz system, the norms of the lag matrices in the ERDMD model span ten orders of magnitude.  Moreover, while ${\bf K}_{1}$ and ${\bf K}_{3}$ have relatively large norms, the remaining lag matrices are $10^{-10}$ times smaller, so all of the long time lag matrices are categorically minuscule in comparison.  So how can they have any meaningful impact within the model?  

\begin{figure}[!h]
\centering
\includegraphics[width=1.\textwidth]{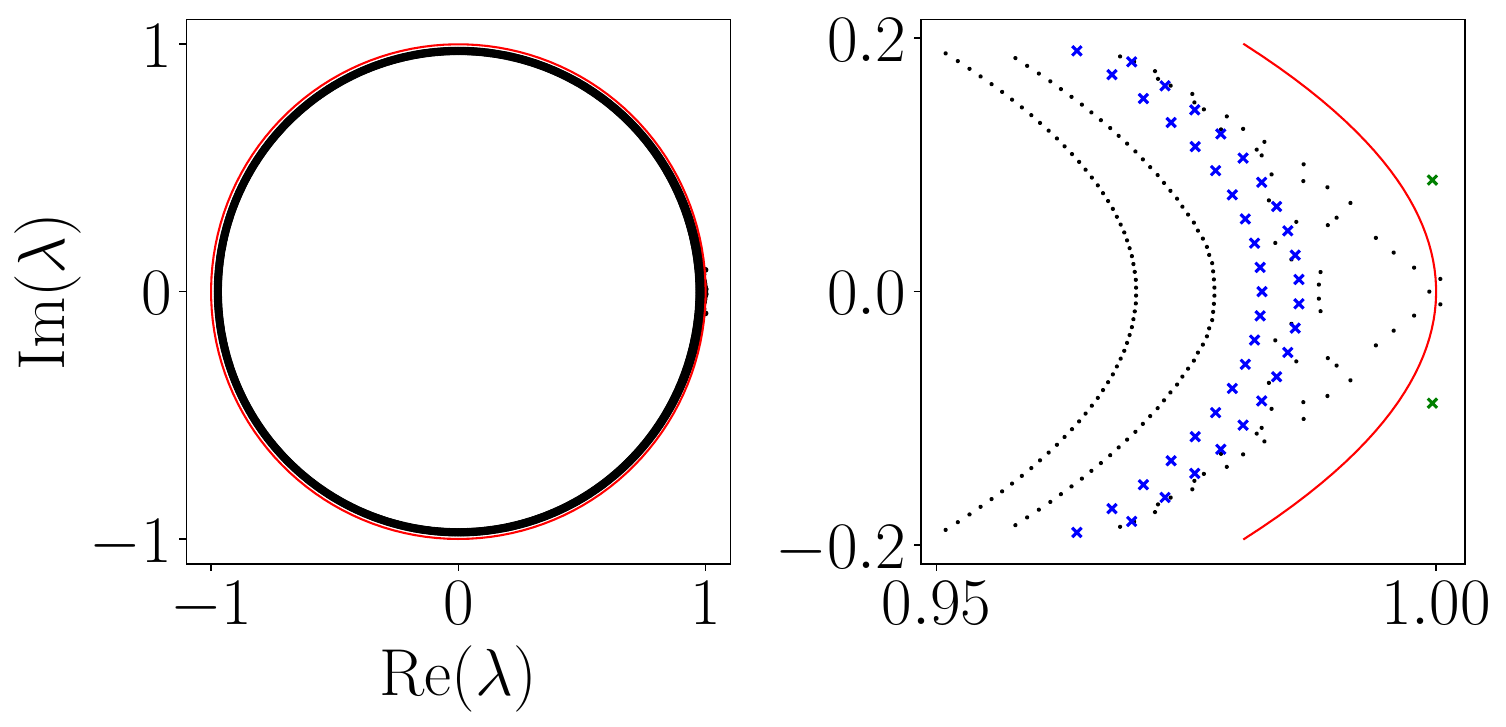}\\
\caption{Spectrum of the corresponding ERDMD Koopman operator for the Rossler system on the left side, with a detail comparison to the roots of $\tilde{p}_{in,a}(z)$ (blue crosses) and $\tilde{p}_{out,a}(z)$ (green crosses) on the right side near (1,0).  The ERDMD algorithm converges to $l_{c}= \left\{1, 3, 170, 436, 553, 665, 988\right\}$.  The solid/red line is the unit circle, provided for comparison.}
\label{fig:rossler_spectrum}
\end{figure}

This question is answered in part by examining the spectrum, which will show us the spread in lag matrix norms is a result of the multiscale nature of the Rossler system.  This of course appears in the wide spread of values of the optimal lags we see in $l_{c}$ and the corresponding spread in norms of the affiliated lag matrices.  Taking $\epsilon=.1$, we see that
\begin{align}
\gnorm{{\bf K}_{1}}, ~ \gnorm{{\bf K}_{3}}=\mathcal{O}\left(\epsilon^{-5}\right),\\ 
\gnorm{{\bf K}_{170}}, ~ \gnorm{{\bf K}_{436}}, ~ \gnorm{{\bf K}_{553}}, ~ \gnorm{{\bf K}_{665}},  ~ \gnorm{{\bf K}_{988}} =\mathcal{O}\left(\epsilon^{5}\right).
\end{align}
Thus, to look at substructure withing $p_{a}(z)$ we should look at the respective balances
\begin{equation}
z^{988}\sim \epsilon^{5}, ~ z^{985}\sim \epsilon^{10}, ~ z^{3}\sim \epsilon^{-5},
\end{equation}
so that for $|z| < \epsilon^{10/985}\approx .977$, we have 
\begin{equation}
p_{a}(z) \approx \tilde{p}_{in,a}(z) = \text{det}\left(z^{323}{\bf K}_{665} + {\bf K}_{988} \right).
\end{equation}
The affiliated eigenvalues computed from this approximation correspond to the slowest scales in our ERDMD model.  See Figure \ref{fig:rossler_spectrum}, right side detail for a comparison of the approximated eigenvalues to the actual values.  

Likewise, if we look for eigenvalues outside the unit circle, we look to the fastest scales corresponding to the reduced polynomial
\[
\tilde{p}_{out,a}(z) = \text{det}\left(z^{3} - {\bf K}_{1}z^{2} - {\bf K}_{3}) \right)
\]
We compare all of these regimes of the spectrum in Figure \ref{fig:rossler_spectrum}, in which we see most eigenvalues are close to or on the unit circle, though near $(1,0)$ as seen in the detail figure, we have a handful of modes outside.  The largest in magnitude of these growing/unstable modes correspond to the fast scale roots coming from $\tilde{p}_{out,a}(z)$.  Likewise, we see in the detail two rings of decaying modes which are reasonably well approximated by $\tilde{p}_{in,a}(z)$, which shows how the relatively minimal model generated by the ERDMD method allows for more ready identification, classification, and approximation of time scale related phenomena.   

\subsection*{Kuramoto--Sivashinsky Equation}

To look at a more intricate and higher dimensional example, we now study the Kuramoto--Sivashinsky (KS) equation given by 
\[
u_{t} + u_{xx} + u_{xxxx} + uu_{x} = 0, ~ u(x+2L,t) = u(x,t).
\]
See \cite{robinson} for an extensive bibliography with regards to details and relevant proofs of facts used in this paper.  Introducing the rescalings
\[
\tilde{t} =\frac{t}{T}, ~ \tilde{x} = \frac{\pi}{L}x, ~ u = A\tilde{u},
\]
and taking the balances
\[
A = \frac{L}{\pi T}, ~ T = \left(\frac{L}{\pi}\right)^{2}, 
\]
we get the equivalent KS equation (dropping tildes for ease of reading)
\[
u_{t} + u_{xx} + \nu u_{xxxx} + uu_{x} = 0, ~ \nu = \left(\frac{\pi}{L}\right)^{2}.
\]
Looking at the linearized dispersion relationship $\omega(k) = k^{2} - \nu k^{4}$, we see that the $\nu$ parameter acts as a viscous damping term.  Thus, as the system size $L$ is increased, the effective viscosity is decreased, thereby allowing for more complex dynamics to emerge.  As is now well known, for $L$ sufficiently large, a fractional-dimensional-strange attractor forms which both produces intricate spatio-temporal dynamics while also allowing for a far simpler representation of said dynamics.  It is has been shown in many different works (see for example \cite{citanovic}) that $L=11$ generates a strange attractor with dimension between eight and nine, and that this is about the smallest value of $L$ which is guaranteed to generate chaotic dynamics.  We therefore set $L=11$ throughout the remainder of this section.  

To study ERDMD on the KS equation, we use KS data numerically generated by a pseudo-spectral in space and fourth-order exponential-differencing Runge-Kutta in time method \cite{kassam} of lines approach.  For the pseudo-spectral method, $K=128$ total modes are used giving an effective spatial mesh width of $2L/K = .172$, while the time step for the Runge-Kutta scheme is set to $\delta t = .25$.  After a burn in time of $t_{b}=10$, we generated a simulation of length $t_{f} = \left(L/\pi\right)^{4}\approx 160$ to allow for nonlinear effects to fully manifest.  This trajectory was then separated via a POD into space and time modes; see \cite{berkooz}.  Taking $N_{s}=12$ modes captured 98.6\% of the total energy.  

\begin{figure}[!h]
\centering
\includegraphics[width=1\textwidth]{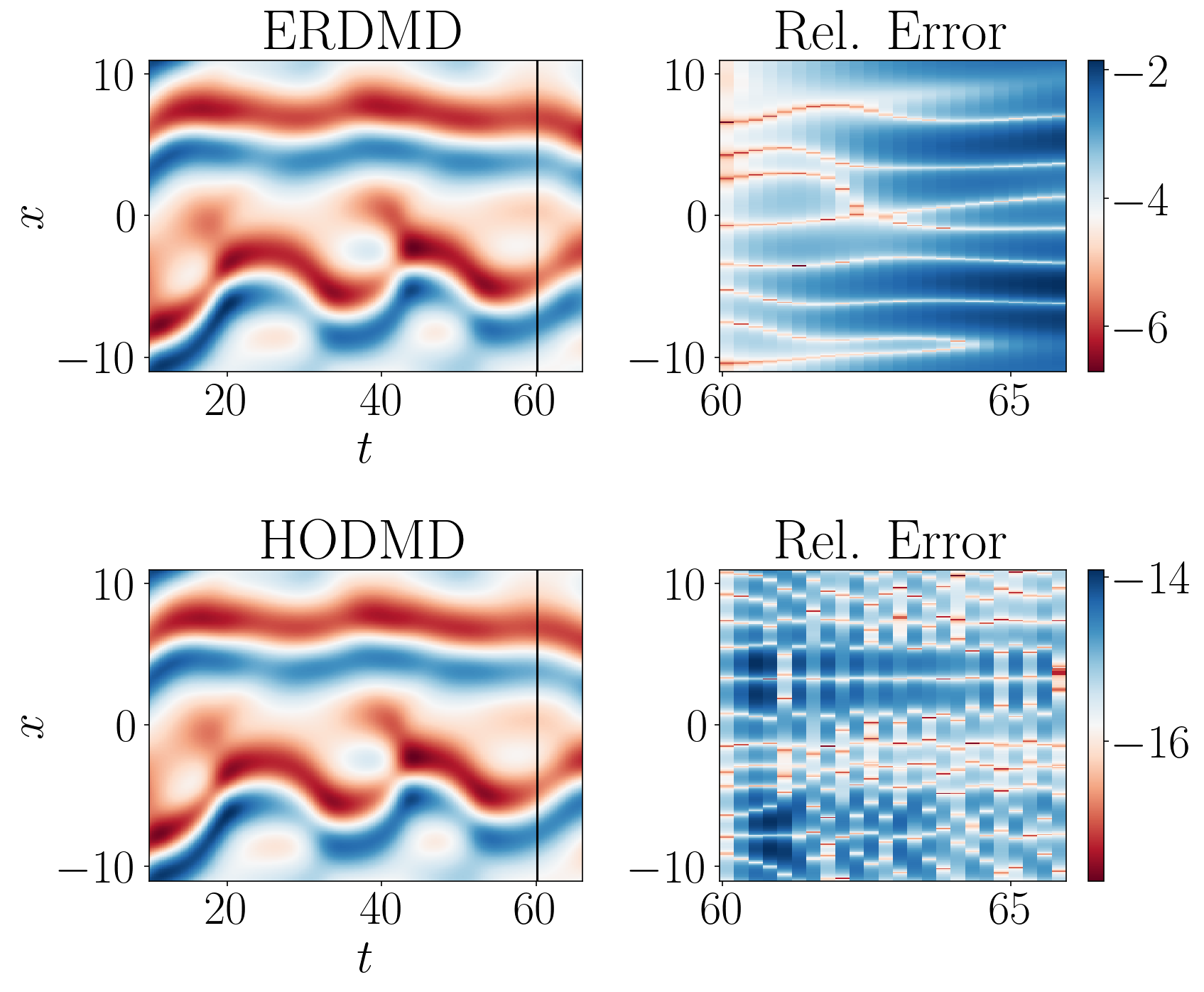}
\caption{Comparison of ERDMD and all lags HODMD model against true trajectory for the KS equation.  The vertical bar indicates the maximum lag choice of $d=200$. The ERDMD algorithm converges to $l_{c}=\left\{1, 123, 141, 158\right\}$.}
\label{fig:ks_compare_d_200}
\end{figure}

To study the ERDMD method, we choose $d=200$, which for $dt=.25$ corresponds to a lag time of $t=60$.  With this choice, the ERDMD method finds $l_{c}$ to be 
\[
l_{c}=\left\{1, 123, 141, 158\right\}.
\]
The results for reconstruction can be seen in Figure \ref{fig:ks_compare_d_200}, we where we look at times $10\leq t \leq 66$.  As can be seen, the ERDMD does well, though we note that if we look for longer reconstructions, errors do start to appear more rapidly for the ERDMD method than the full HODMD method, and we are not able to push the ERDMD method to reproduce the entire reconstruction region from initial conditions.  Thus, we while the present results are promising, they represent an edge for our method that will need further tools to address; see \cite{curtis_dldmd} in this direction.    
 
\begin{figure}[!h]
\centering
\includegraphics[width=.7\textwidth]{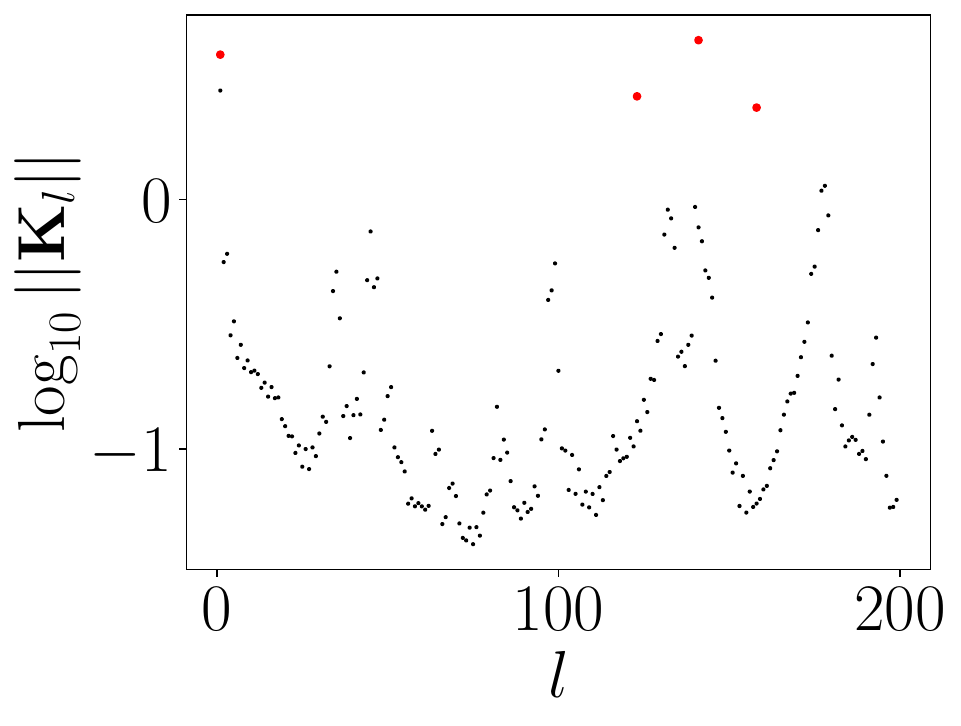}
\caption{Comparison of the HODMD lagged matrix norms (black dots) and the ERDMD model (red dots) for the Kuramoto--Sivashinsky system.}
\label{fig:model_comp_d_200}
\end{figure}

Repeating the HODMD and ERDMD model comparisons, we see in Figure \ref{fig:model_comp_d_200} a situation similar to that seen in the Lorenz-63 system in so far as each matrix within our model is of roughly equal size and thus impact.  Comparing to the Rossler system results, this shows there is no rapid transition layer or other distinguished time scale phenomena.  
 
\begin{figure}[!h]
\centering
\begin{tabular}{c}
\includegraphics[width=1\textwidth]{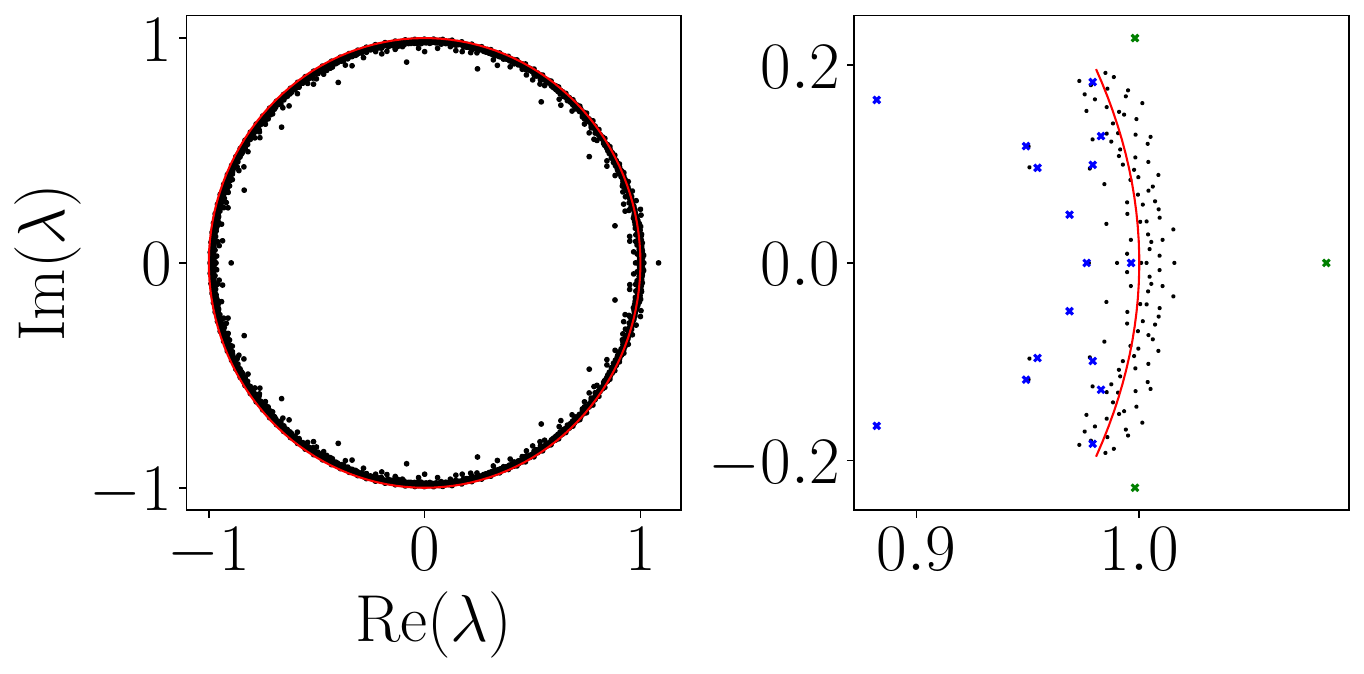} \end{tabular}
\caption{Spectrum of the corresponding ERDMD Koopman operator for the Kuramoto--Sivashinsky system on the left side, with a detail comparison to the roots of $\tilde{p}_{in,a}(z)$ (blue crosses) and $\tilde{p}_{out,a}(z)$ (green crosses) on the right side near (1,0).  The ERDMD algorithm converges to $l_{c}=\left\{1, 123, 141, 158\right\}$.  The solid/red line is the unit circle, which is provided for reference.}
\label{fig:ksspectrum}
\end{figure}

Nevertheless, we can still identify inner and outer approximations to eigenvalues affiliated with the ERDMD method.  We see in Figure \ref{fig:ksspectrum} the corresponding full spectrum affiliated with our method.  Again, the strong separation of lags in $l_{c}$ motivate looking at the roots of the reduced polynomial
\[
\tilde{p}_{in, a}(z) = \text{det}\left(z^{35}{\bf K}_{123} + z^{17}{\bf K}_{141} + {\bf K}_{158} \right).
\] 
Note, letting $\tilde{z}=z^{17}$, we can write 
\[
z^{35} = \tilde{z}^{2}e^{\ln \tilde{z}/17} \approx \tilde{z}^{2}
\]
so long as $|z|$ is not too close to zero.  We can then find the affiliated approxmiate companion matrix representation for $\tilde{p}_{a}(z)$ in the form
\[
\tilde{{\bf K}}_{a} = 
\begin{pmatrix} 
0 & I \\ 
-\tilde{{\bf K}}_{2} & -\tilde{{\bf K}}_{1} \\
\end{pmatrix}
\]
where 
\[
\tilde{{\bf K}}_{1} = {\bf K}_{123}^{-1}{\bf K}_{141}, ~ \tilde{{\bf K}}_{2} = {\bf K}_{123}^{-1}{\bf K}_{158}.
\]
For this problem, for $|z|>1$, we can also readily define and compute the roots of $\tilde{p}_{out,a}(z)$ where
\[
\tilde{p}_{out,a}(z) = \text{det}(z - {\bf K}_{1})
\]

The full spectrum and detail comparison is seen in Figure \ref{fig:ksspectrum}, where we again can see the most damped modes come from the highest lags in our Koopman model.  We likewise see that the most unstable mode of the full spectrum correspond to the largest magnitude eigenvalues of ${\bf K}_{1}$.  Thus there is a clear separation in the Koopman spectrum across lag values, which is to say time scales in the dynamics.     

\section{Discussion}
In this work, we have developed a novel lagged DMD method which uses entropic regression to discover relatively minimal, non-uniform in lag models which are shown to allow for direct time stepping with good accuracy.  Further, the non-uniform lag structure and relatively small number of model terms used allow for more detailed study and characterization of the affiliated DMD spectrum, thereby adding diagnostic depth which should prove useful in a number of real-world contexts.  While we also discovered limitations in our approach, future work involving existing machine learning based DMD algorithms should provide ready improvement.  Finally of course, there is the frontier of studying our method on noisy data, though existing work of one of the present authors hints that entropic based regression approaches should be able to address that issue well.  

\section*{Acknowledgments}
Curtis and Lago would like to acknowledge the support of ONR grant N00014-23-1-2106.

\bibliographystyle{unsrt}
\bibliography{ionosphere_bib}
\end{document}